\documentclass[sigconf]{aamas} 

\usepackage{balance} 
\usepackage{array, multirow}
\usepackage{graphicx}
\usepackage{latexsym}
\usepackage{fancyvrb}
\usepackage{amsmath} 
\usepackage{mathtools}
\usepackage{centernot}
\usepackage{algpseudocode}
\usepackage{algorithm}
\usepackage{xcolor}
\usepackage{soul}
\usepackage{tikz}
\usetikzlibrary{trees}
\usepackage{caption,subcaption}
\usepackage{bm}
\usepackage{url}
\usepackage{paralist}
\usepackage{hyperref}

\setcopyright{ifaamas}
\acmConference[AAMAS '23]{Proc.\@ of the 22nd International Conference
on Autonomous Agents and Multiagent Systems (AAMAS 2023)}{May 29 -- June 2, 2023}
{London, United Kingdom}{A.~Ricci, W.~Yeoh, N.~Agmon, B.~An (eds.)}
\copyrightyear{2023}
\acmYear{2023}
\acmDOI{}
\acmPrice{}
\acmISBN{}

\acmSubmissionID{944}

\title[Predicting PrivacyPreferences for Smart Devices as Norms]{Predicting Privacy Preferences for Smart Devices as Norms}

\author{Marc Serramia}
\affiliation{
  \institution{King's College London}
  \city{London}
  \country{United Kingdom}}
\email{marc.serramia_amoros@kcl.ac.uk}

\author{William Seymour}
\affiliation{
  \institution{King's College London}
  \city{London}
  \country{United Kingdom}}
\email{william.1.seymour@kcl.ac.uk}

\author{Natalia Criado}
\affiliation{
  \institution{Universitat Politècnica de València}
  \city{València}
  \country{Spain}}
\email{ncriado@upv.es}

\author{Michael Luck}
\affiliation{
  \institution{King's College London}
  \city{London}
  \country{United Kingdom}}
\email{michael.luck@kcl.ac.uk}

\begin{abstract}
Smart devices, such as smart speakers, are becoming ubiquitous, and users expect these devices to act in accordance with their preferences. In particular, since these devices gather and manage personal data, users expect them to adhere to their privacy preferences. However, the current approach of gathering these preferences consists in asking the users directly, which usually triggers automatic responses failing to capture their true preferences. In response, in this paper we present a collaborative filtering approach to predict user preferences as norms. These preference predictions can be readily adopted or can serve to assist users in determining their own preferences. Using a dataset of privacy preferences of smart assistant users, we test the accuracy of our predictions.
\end{abstract}



\newtheorem{mydef}{Def.}

\newtheorem{myexample}{Example}

\newtheorem*{discussion*}{Discussion}
\newtheorem*{encoding*}{Encoding}

\newtheorem*{observation*}{Observation}
\newcommand{\marc}[1]{\textcolor{black}{#1}}

\begin{document}


\pagestyle{fancy}


\maketitle 


\section{Introduction}\label{sec:intro}

Artificial intelligence (AI) technologies are making their way into our daily lives and into our homes. We have grown accustomed to using our devices 
to call friends, set reminders, or check the weather.
However, for these technologies to be adopted and trusted by users, they must act as users expect, and this problem is especially apparent in the area of privacy preferences. Studies show that users are deeply concerned about how their data is being collected online \cite{madden2014perceptions}. Interestingly, while they expect AI to act as they desire, they are unwilling to spend time setting their preferences. 
For example, despite users' concerns about privacy, studies show that they ignore or blindly accept cookie banners \cite{kretschmer2021cookie} and privacy policies in social networks \cite{Obar2018lie}. Furthermore, in social networks, a large proportion of users do not change default privacy settings \cite{Krishnamurthy2009leakage}. This can be explained as a result of privacy fatigue \cite{Choi2018Fatigue}, the sensation of loss of control and futility over protecting one's privacy. This leads to privacy cynicism, when users do not adopt a privacy protecting behaviour even if they are concerned about their privacy \cite{Hoffmann2016Cynicism}.
Thus, the current approach of directly asking the user when a preference is unknown but needed fails to capture the user's true preferences. Additionally, continual questioning prevents users from achieving their objectives with the device. In response, this paper advocates for an approach that can understand user preferences with less user involvement, in turn bringing more importance to user interactions whenever such preferences are needed.

A particular platform in which capturing privacy preferences is challenging and yet essential is that of smart speakers and other smart personal assistants. These devices have benefited from widespread early adoption, and it is  estimated that 500 million units were installed in the last quarter of 2021~\cite{Strategy2021smartspeakerinstalled}. Nonetheless, the early adoption of these technologies means that they still have several vulnerabilities that pose a threat to the security and privacy of their users \cite{EduSPASurvey}. Indeed, there have already been cases reported in which smart assistants have not functioned as expected; for example, 
a smart speaker recorded and sent a private conversation without the user's consent~\cite{GuardianAlexa}. These situations hinder user trust in the technology and can ultimately lead users to limit the functionalities of the devices used, or even to adopting coping mechanisms \cite{Abdi2019More}.

This paper describes an alternative approach that addresses the issues outlined above. The critical observation underpinning our approach is that smart devices are just one part of a larger ecosystem (e.g. see \cite{EduSPASurvey} for a description of the ecosystem of smart speakers), and they interact and share data with agents like services, apps, and other devices. For example, a smart watch might send a voice recording to a smart speaker, or might share the wearer's heart rate with a health app. In this respect, we can understand this ecosystem as a multi-agent system in which the use of norms can help to regulate these interactions, implementing privacy preferences. 

Norms can effectively summarise complex privacy preferences into simple sets of regulations, as shown by Abdi et al. \cite{NouraPrivacyNorms2021}, who gathered over 800 privacy preferences on data transmissions, yet produced just 17 norms. Moreover, although here we assume no knowledge of the domain, if such knowledge is available there exist techniques to generalise norms (see \cite{MoralesAAMAS2013, MoralesAAMAS2014} for an example) or find and resolve inconsistencies among them \cite{VasconcelosKN09}. Furthermore, norms are also used by people, and are naturally understood by them, representing a good base upon which to construct explanations. This can be used not only to generate explanations for a user if something unexpected happens, but also to tailor interactions with a user to validate predicted norms. 
Norms are regarded as expected patterns of behaviour \cite{WooldridgeIntrodMAS}, causing agents (each component in the smart device ecosystem) to coordinate better and function more efficiently. As an example with smart devices, imagine a service knows in advance the privacy norms of a user with regard to each component of the ecosystem. If this service needs to interact with other components, it can use the user's norms to adapt its behaviour to avoid violating norms or to avoid performing unregulated transmissions of information, which might require 
consent.

\marc{As informally outlined in \cite{SerramiaPRIMA22Collaborative}, we can} exploit the large user bases of smart devices to use knowledge of previously specified privacy preferences to infer new preferences or to assist users in specifying their preferences. In particular, we aim to exploit similarities between users to make privacy preference predictions using collaborative filtering \cite{Su2009Collaborative}. \marc{Effectively, we see the smart device ecosystem as a multi-layered multi-agent system. The lower level represents the multi-agent system associated with a single user (that is, the user's device, and the other devices, skills, and services that can be accessed from it). The higher level is that of the multi-agent system composed of all the users. Our approach is centred on the norm creation stage in the lower level multi-agent systems related to each user. Therefore, each device user has its own associated set of norms, and all agents within its lower level multi-agent system, be they devices, skills, or other services, are informed and affected by the norms whenever they want access to the user's personal data.}
While many researchers have studied different approaches to constructing norm systems, like norm synthesis \cite{MoralesAAMAS2013, morales2015compact} or norm emergence \cite{shoham1997emergence, SavarimuthuAAMAS07, sugawara2011emergence},
we are not aware of any similar approach like the collaborative filtering presented here.

In taking this approach, we make the following contributions.
\begin{itemize}
    \item Formalisation of the problem of predicting norms to ensure that computational behaviour aligns with user preferences. This is divided into two subproblems, namely preference approximation (predicting unknown user preferences) and norm inference (inferring norms from predicted preferences).
    \item Formalisation of preference prediction functions. We provide a specific example of this type of function based on the preferences of similar users.
    \item Inference of norms from the predicted preferences, and specification of different methods to do so based on the confidence of the prediction or other variables.
\end{itemize}
The paper is structured as follows: Section \ref{sec:definition} formalises the core problems we aim to address in the paper. In Section \ref{sec:prediction} we detail the process of predicting preferences. We then use these predictions to infer norms in Section \ref{sec:inference}. Section \ref{sec:case} is dedicated to validate our findings. In Section \ref{sec:rw} we discuss related work. Finally, in Section \ref{sec:conc} we discuss conclusions and future work.
\section{Problem definition}
\label{sec:definition}

Consider a set of users $U$ and a set of agents $Ag$, such that $ag_{u} \in Ag$ is the agent (i.e.\ the smart device) of $u \in U$\footnote{To simplify, and without loss of generality we assume that for each $u \in U$ there is only one $ag_{u} \in Ag$. Note that if one user had more than one device, we could consider a mock second user.}. Consider also a finite number of elements $X = \{x_1, \dots, x_{|X|}\}$ over which users have preferences. These elements will commonly be actions an agent can perform, but can also be more complex, for example containing the context in which an action happens \marc{(e.g. ``share if the user is notified'')}. For generality purposes, we do not specify the formalisation of these elements since, for the problem definition (and our proposed resolution), it is not necessary. \marc{This not only allows our notation to be kept simple, but it also allows us to define the preference domain with as much or as little complexity as needed.} Given an element $x \in X$, we assume the user's preference towards $x$ is a number in $[-1, 1]$, where 1 means the user totally approves of $x$, -1 means the user totally disapproves of $x$, and 0 means neutrality towards $x$. Note that we can make this assumption without loss of generality as we can always transform any user preferences into $[-1, 1]$\footnote{On the one hand, numerical preferences can be re-scaled into $[-1, 1]$, because the number of users and the number of elements is finite and therefore preferences will always be bounded. On the other hand, ordinal preferences will also be bounded and can be transformed into numerical preferences in $[-1, 1]$.} 

For each user $u$ and agent $ag_{u}$, we consider the following tuples of preferences.

\begin{itemize}
    \item The user's preference profile $p_{u}$ represents the real preferences of user $u \in U$. Note that $p_{u} \in [-1, 1]^{|X|}$, and the $i^{th}$ position in the tuple represents the preference of user $u$ towards element $x_i$.
    \item The agent's preference profile $p_{ag_{u}}$ represents the preferences of user $u$ known by agent $ag_{u}$. This tuple has the same structure as $p_{u}$, but unlike $p_{u}$, this tuple has gaps of knowledge. We represent an unknown preference as $\bigcirc$, therefore $p_{ag_{u}} \in ([-1, 1]\cup\{\bigcirc \})^{|X|}$.
\end{itemize}

Having introduced these elements, we now present a running example with smart personal assistants, which we use throughout the paper to illustrate the concepts we introduce. 

\begin{myexample}
\label{ex:formal}
We consider users $u_1, u_2$ and $u_3$, who have smart personal assistants $ag_{u_1}, ag_{u_2}$ and $ag_{u_3}$ respectively. We consider three elements over which users have preferences: sharing data with the AI assistant manufacturer ($x_1$), 
with internet provider ($x_2$), and 
with developers of third party skills ($x_3$). When it comes to the user's real preferences, we have: $p_{u_1} = (-1, -1, -1)$, $p_{u_2} = (-1, -1, -1)$, $p_{u_3} = (1, -1, 1)$.
As for the agent's known preferences, we have: $p_{ag_{u_1}} = (-1, -1, \bigcirc)$, $p_{ag_{u_2}} = (-1, \bigcirc, -1)$, $p_{ag_{u_3}} = (1, \bigcirc, 1)$.
\end{myexample}

Finally, we define a process to complete preferences, noted as $comp$. This process takes $p_{ag_{u}}$ and completes it, producing $p_{ag_{u}}^* \in [-1, 1]^{|X|}$. Note that in $[-1, 1]^{|X|}$ we can assess distances between preference tuples (which are points in the space). With this in mind, we can formalise the first problem we address in this paper.

\begin{mydef}[Preference approximation problem]
Consider the space $[-1, 1]^{|X|}$, and $dis$ a distance function in this space, the preference approximation problem consists of finding the process $comp$, with the aim of minimising the distance $dis(p_{u}, p_{ag_{u}}^*)$.
\end{mydef}

\begin{myexample}
The process to complete preferences could be, for example, asking the users about their preferences directly. For example, user $u_1$ interacts with a third party skill that requires unknown preferences and therefore asks the user about them. The user might want to use the skill and responds affirmatively automatically (against their real preferences), thus $p_{ag_{u_1}}^* = (-1, -1, 1)$. Then, considering the euclidean distance, we would have $dis(p_{u_1}, p_{ag_{u_1}}^*) = 2$, which means that, in this case, this process can be improved.
\end{myexample}

Our final aim is to align agent behaviour to user preferences. To that end, we resort to norms to regulate how each agent $ag_{u}$ behaves. Note that there is no standard definition of norm; for example, \cite{LopezLI02} considers rewards and punishments in norms, whereas \cite{morales2015compact} ignores these and instead considers the context of application of the norm. In our case, we use a very simple definition of norm in support of generality, as a more complex definition would require domain knowledge which would hinder the applicability of our approach. 

\begin{mydef}[Norm]
Given an element $x \in X$, a norm is a structure $\theta(x)$, where $\theta \in \{Prh, Per\}$, where $Prh(x)$ is the norm prohibiting $x$, whereas $Per(x)$ is the norm permitting $x$.
\end{mydef}

Having defined our notion of norm, we can now define the second problem we address in this paper, as follows: 

\begin{mydef}[Norm inference problem]
Given an agent $ag_{u} \in Ag$ and its completed preferences $p_{ag_{u}}^*$, the norm inference problem consists of enacting preferences in $p_{ag_{u}}^*$ as norms, such that when following them, the agent will behave as expected by user $u$.
\end{mydef}

\begin{myexample}
Supposing we correctly completed the preferences of user $u_3$ ($p_{ag_{u}}^* = (1, -1, 1)$), our aim would be to find a way to encode these preferences as norms, those close to 1 into permission norms, and those close to -1 into prohibition norms. In this case, $Per(x_1)$, $Prh(x_2)$, and $Per(x_3)$.
\end{myexample}
\section{Preference prediction}
\label{sec:prediction}

In this section we consider how to predict a user's preference toward an element $x$ for which we don't know their preference. To do so, we will infer preferences from similar users. As argued above, we can assume that users who share similar views on known preferences will also share similar views on unknown ones. Our aim is to formally define a separation measure between users so that we can build predictions based on a set of users deemed similar enough. This can be an aggregation of the preferences of similar users over the element in question. First, however, we must introduce some preliminary notation and definitions and, to simplify, we reuse the notation introduced above. Given a user $u \in U$ with corresponding agent $ag_{u}$ and an element $x \in X$, we note the real preference of $u$ towards $x$ as $p_{u}(x)$, the known preference of $u$ towards $x$ by agent $ag_u$ as $p_{ag_{u}}(x)$, and the preference of $u$ towards $x$ by agent $ag_u$ after the prediction process as $p*_{ag_{u}}(x)$.
First, we define common known preference elements. Given a pair of users, the common known preference elements are those elements for which we know both users' preferences, and can be used to measure their separation.

\begin{mydef}[Common known preference elements]
Consider two users $u_1, u_2$, their agents $ag_{u_1}, ag_{u_2}$, and the users' known preferences $p_{ag_{u_1}}$ and $p_{ag_{u_2}}$. We call the common known preference elements of $ag_{u_1}$ and $ag_{u_2}$, the set of elements for which we know both agents' preferences. If $p_{ag_{u_1}} = (p_1^1, \dots, p_1^{|X|})$ and $p_{ag_{u_2}} = (p_2^1, \dots, p_2^{|X|})$, this is formalised as:
$$
C(ag_{u_1}, ag_{u_2}) = \{x_i | p_{u_1}(x_i), p_{u_2}(x_i) \neq \bigcirc\}
$$
\end{mydef}

With these preliminary definitions, we now turn to formalising a measure of separation between users. While we can define distance functions in the space of real user preferences $[-1, 1]^{|X|}$, our aim here is to assess distances between users with only partial knowledge of their preferences. Therefore, we want to define a distance in the space of known preference tuples $([-1, 1]\cup\{\bigcirc\})^{|X|}$, though in this case we do not want a strict distance function, but a more relaxed version of a distance function. 
Consider, for example, the points $(1,\bigcirc, \dots, \bigcirc)$ and $(\bigcirc, \dots, \bigcirc, 1)$, where a distance function would have to assign a distance between these two points, but there are no commonly known preferences between them, so we choose not to assign a distance in this case. In other words, the separation between two users should only depend on their commonly known preferences. Hence, instead of defining a formal distance we define a function, called a preference separation function, for which we require similar properties to those of distances, albeit more relaxed.

\begin{mydef}[Preference separation]
\label{def:sep}
Given a set of pairs of users with common elements $U_{com} = \{(u, u') \in U \times U | C(u, u') \neq \emptyset\}$, a user separation measure is a function $sep: U_{com} \rightarrow \mathbb{R}$ that measures the separation of two users $(u_1, u_2) \in U_{com}$, based on their known preferences by the agent (i.e. $p_{ag_{u_1}}$ and $p_{ag_{u_2}}$). This function must satisfy the following properties:
\begin{itemize}
    \item \textbf{Dependence of commonly known preferences:} $sep(u_1, u_2)$ only depends of $p_{ag_{u_1}}(x)$ and $p_{ag_{u_2}}(x)$, $\forall x \in C(ag_{u_1}, ag_{u_2})$
    \item \textbf{No-negativity:} $sep(u_1, u_2) \geq 0$.
    \item \textbf{Symmetry:} $sep(u_1, u_2) = sep(u_2, u_1)$.
    \item \textbf{Zero separation for equal known common preferences:} $sep(u_1, u_2) = 0 \Leftrightarrow p_{ag_{u_1}}(x) = p_{ag_{u_2}}(x) \forall x \in C(ag_{u_1}, ag_{u_2})$.
    \item \textbf{Triangle inequality for known common preferences:} Given a third user $u_3$, if we note $C = C(ag_{u_1}, ag_{u_2})$, then $sep(u_1, u_2) \leq sep{\big|}_{C}(u_1, u_3) + sep{\big|}_{C}(u_3, u_2)$. where $sep{\big|}_{C}$ is the separation function restricted to only the common elements of $ag_{u_1}$ and $ag_{u_2}$ (i.e. applying $sep$ as if $X = C$).
\end{itemize}
\end{mydef}

As argued earlier, the first property ensures that user separation only depends on the commonly known preferences of the users. The next four properties correspond to the properties of distances adapted to our case, as follows. First, we require that the separation measure is positive, as 0 is the closest possible separation, disallowing negative separations. Second, this function must be symmetric as the separation between two preferences should be the same no matter the order. Third, two users have separation of 0 if and only if their commonly known preferences are the same. This property is more general than $sep(u_1, u_2) = 0 \Leftrightarrow u_1 = u_2$ as we do not want to take into account what happens with not commonly known preferences. The fourth property is a general version of the triangle inequality where we only consider the commonly known preferences. Again, we want to be more general because we want to disregard preferences for which we do not have complete knowledge of both users.
To illustrate,
we provide an example of such a function, which we call cumulative user separation.

\begin{mydef}[Cumulative user separation]
\label{def:exsep}
The cumulative user separation function is a function $sep_{+}: U_{com} \rightarrow \mathbb{R}$ that, for users $(u_1, u_2) \in U_{com}$ and respective agents $ag_{u_1}, ag_{u_2}$, assesses their separation as follows:
$$
sep_{+}(u_1, u_2) = \sum_{x_i \in C(ag_{u_1}, ag_{u_2})} |p_{ag_{u_1}}(x_i) - p_{ag_{u_2}}(x_i)|
$$
\end{mydef}

Proving that $sep_{+}$ satisfies the properties in Definition \ref{def:sep} is straightforward, but omitted due to space constraints.

The following example illustrates the concepts introduced so far, in context of the scenario introduced in Example \ref{ex:formal}.

\begin{myexample}
\label{ex:sep}
We assess the separation between $u_1$ and the other users. For both $u_2$ and $u_3$, we have $C(ag_{u_1}, ag_{u_2}) = C(ag_{u_1}, ag_{u_3}) = \{x_1\}$, as for $u_1$ we know preferences over $x_1$ and $x_2$, but for the other users we know preferences over $x_1$ and $x_3$, but not $x_2$. Thus, in this case the cumulative separation is $sep_{+}(u_1, u_2) = 0$ and $sep_{+}(u_1, u_3) = 2$.
\end{myexample}

Given a user for which we want to predict a preference over $x$, we can gather a set of similar users for which we know their preferences over $x$ considering their separation with regard to the original user. With this set of similar users, we can then predict the targeted preference by aggregating the preferences of similar users towards that element. With this aim, we define the set of similar users. Since we build this set to make predictions, we must require that we know the preference of the users in the set with regard to the targeted element. Ideally, similar users should be those with separations less or equal to a maximum $\epsilon$. However, since our aim is to use similar users to build predictions, we require a minimum number of similar users $\nu$ (those with the least separation), so that predictions are founded on a reasonable number of users.

\begin{mydef}[$\epsilon\nu$-similar users]
\label{def:ensimilar}
Given a user $u$, an element $x$, and parameters $\nu \in \mathbb{N}$ and $\epsilon \in \mathbb{R}$, we call $\epsilon\nu$-similar users the set $Sim^\epsilon_\nu(u, x)$ of similar users to $u$, such that they have preferences over $x$, and which contains at least the $\nu$ most similar users and all users that are closer than $\epsilon$ (in terms of separation). Hence, if $K(x) = \{u| p_{ag_{u}}(x) \neq \bigcirc\}$ is the set of users for whom we know their preference over $x$, we formalise $Sim^\epsilon_\nu(u, x) = Sim^{\epsilon}(u, x) \cup Sim_{\nu}(u, x)$, where:
\begin{itemize}
\item $Sim^{\epsilon}(u, x) = \{u'\in K(x) | sep(u, u') \leq \epsilon\}$
is the set of users with known preference over $x$ who have a separation with $u$ less or equal to $\epsilon$.
\item $Sim_{\nu}(u, x) = \{u_1, \dots, u_{\nu} \in K(x) | sep(u, u_1) \leq \dots \leq sep(u, u_{\nu})\allowbreak\text{ and }\nexists u' \in K(x)\setminus \{u_1, \dots, u_{\nu}\}\allowbreak \text{s.t. } sep(u, u') < sep(u, u_{\nu})\}$ is the set of the $\nu$ closer users to $u$ with known preference over $x$.
\end{itemize}
\end{mydef}

Using the set of similar users, we can predict the targeted preference, for which we use a prediction function, defined as follows.

\begin{mydef}[Preference prediction function]
A preference prediction function is a function $pre: U \times X \rightarrow [-1, 1]$ which, given a pair of a user $u\in U$ and elements $x \in X$, predicts the preferences of $u$ towards $x$. Given $\nu \in \mathbb{N}$ and $\epsilon \in \mathbb{R}$, this function must depend only on the preferences towards $x$ of similar users to $u$, $\{p_{ag_{u'}}(x) | u' \in Sim^\epsilon_\nu(u, x)\}$
\end{mydef}

We provide an example of a preference prediction function called average preference prediction function. This function builds a prediction of the preference of $u$ towards $x$ as the average of the preferences over $x$ of similar users to $u$. Formally:

\begin{mydef}[Average preference prediction function]
\label{def:avgpred}
Given a separation measure $sep$, the average preference prediction function $pre_{avg}: U \times X \rightarrow [-1, 1]$, takes a user $u\in U$ (with $p_{ag_{u}}(x) = \bigcirc$) and an element $x_i\in X$, and predicts the preference of $u$ towards $x$ in $[-1, 1]$, as follows:
$$
pre_{avg}(u, x) = \frac{\sum_{u' \in Sim^\epsilon_\nu(u, x)} p_{ag_{u'}}(x)}{|Sim^\epsilon_\nu(u, x)|}
$$
\end{mydef}

We continue with the following illustration of a prediction.

\begin{myexample}
\label{ex:pred}
We want to predict the preference of $u_1$ with regard to $x_3$. Considering the separations found in Example \ref{ex:sep}, we aim at selecting the similar users. To do so, we require at least one user (i.e. $\nu = 1$) and we consider them similar if the separation is less than 0.5 (i.e. $\epsilon = 0.5)$, then $Sim^{\epsilon = 0.5}_{\nu = 1}(u_1, x_3) = \{u_2\}$. Then, to predict the preference of $u_1$ towards $x_3$, we average the preferences of the similar users towards $x_3$. In the case of Example~\ref{ex:sep}, the result would be $pre_{avg}(u_1, x_3) = -1$
\end{myexample}

Given a prediction function, we can complete the unknown preferences of the user as follows.

\begin{mydef}[Complete predicted preferences]
\label{def:precomp}
Given a preference prediction function $pre$ and user $u$ with known partial preferences $p_{ag_{u}}$, the tuple of complete predicted preferences for the user is $p*_{ag_{u}} = (p*_{ag_{u}}^1, \dots, p*_{ag_{u}}^{|X|})$, composed of the known preferences and predictions of the unknown ones. Formally:
$$
p*_{ag_{u}}^i = 
\left\{
\begin{array}{ll}
      p_{ag_{u}}^i & \text{ if } p_{ag_{u}}^i \neq \bigcirc \\
      pre(u, x_i) & \text{ if } p_{ag_{u}}^i = \bigcirc
\end{array} 
\right.
$$
\end{mydef}

Note that in Example \ref{ex:pred}, the preference predicted along with the known preferences form the complete predicted preferences for $u_1$.

Using Definition \ref{def:precomp}, we obtain the complete preferences of the user from the already known preferences and the newly predicted ones. This offers a solution to the preference approximation problem, and we show the validity of our approach in Section \ref{sec:case}. First, however, we tackle the norm inference problem in the next section.





\section{Norm Inference from predictions}
\label{sec:inference}

At this point, we can predict user preferences from similar users. However, our broader aim is to build norms from these preferences so that agents can follow them. This means transforming numerical preferences in [-1, 1] into norms. In this section, we propose several methods to perform this transformation and discuss under which circumstances these methods would be appropriate to be used.

\subsection{Hard thresholds}
\label{sub:hard}

The simplest method we can use to transform numbers in [-1, 1] into norms is through hard thresholds. Thus, we would consider two thresholds $\epsilon_{prh}$, and $\epsilon_{per}$ that divide [-1, 1] into three blocks, referring to (in the following order): prohibition, no norm, and permission. Hence, for consistency, we require that the threshold of prohibition must be on the negative side of the preferences interval, and the permission threshold on the positive side, $\epsilon_{prh} \in [-1, 0]$, and $\epsilon_{per} \in [0, 1]$. Then, considering the completed preferences $p*_{ag_{u}}$, we would build norm $Prh(x)$ if $p*_{ag_{u}}(x) \leq \epsilon_{prh}$, no norm if $\epsilon_{prh} < p*_{ag_{u}}(x) < \epsilon_{per}$, or $Per(x)$ if $\epsilon_{per} \leq p*_{ag_{u}}(x)$.

\begin{myexample}
If we have thresholds $\epsilon_{prh} = -0.25$ and $\epsilon_{per} = 0.25$, then elements $x$ with preferences in $[-1, -0.25]$ would be prohibited ($Prh(x)$), those in $[-0.25, 0.25]$ would not be regulated, and those in $[0.25, 1]$ would be permitted ($Per(x)$).
\end{myexample}

\subsection{Thresholds based on prediction confidence}

Note that hard thresholds can be problematic when predictions are not particularly accurate (for example, due to $p_{ag_{u}}$ having many unknown preferences). In this case, for unknown preferences that are close to the threshold our prediction can easily fall on either side. Thus, in these cases we can consider variable thresholds depending on the confidence of our predictions. 

Here, we consider thresholds to be a function of prediction confidence. If we consider confidence to be a number in $[0, 1]$, then we formalise thresholds as functions: $\epsilon_{prh}: [0, 1] \rightarrow [-1, 0]$ and $\epsilon_{per}: [0, 1] \rightarrow [0, 1]$.

The remaining task now is to define prediction confidence. Note that we would consider a prediction based on other very similar agents, with very similar preferences, as being accurate, whereas a prediction obtained from agents close in opinion but not entirely similar, and whose preferences span over an array of options, likely not very accurate. If we do not know the real preference we cannot be entirely sure of the quality of predictions but confidence gives us an intuition on the quality of the data they are drawn from. Formally, we define a prediction confidence function as follows.

\begin{mydef}[Confidence function]
\label{def:confidence}
A prediction confidence function $conf: U \times X \rightarrow [0, 1]$ is a function that takes a pair of user and element and gives the confidence of prediction $pre(u,x)$ in $[0,1]$, where 0 means no confidence and 1 is absolute confidence. Note that in general $conf(u, x) > conf(u', x')$ should imply $|pre(u,x) - p_{u}(x)| < |pre(u',x') - p_{u'}(x')|$. In other words, a higher confidence should correlate with a better prediction (one closer to the real preference).
\end{mydef}

We provide an example prediction confidence function called $\rho\mu$-Confidence based on the following two measures: 
\begin{itemize}
\item The separation between $u$ and users in $Sim^\epsilon_\nu(u, x)$ (for some separation measure $sep$)
\item The distribution of preferences of users in $Sim^\epsilon_\nu(u, x)$ towards $x$ (i.e. their standard deviation).
\end{itemize}

We define $\rho\mu$-Confidence as the weighted average of these two measures where $\rho$ and $\mu$ are the weights. 

\begin{mydef}[$\rho\mu$-Confidence]
\label{def:rmconf}
Let $sep$ be a separation measure as in Def.~\ref{def:sep} and $Sim^\epsilon_\nu(u, x)$ be the set of similar users to user $u$ (using $sep$). We can then define the confidence of prediction $p(u, x)$ as:
$$
conf_{\rho, \mu}(u, x) = 1- \rho\cdot\min(\frac{\sum_{u' \in Sim^\epsilon_\nu(u, x)}sep(u, u')}{|Sim^\epsilon_\nu(u, x)|}, 1) - \mu\cdot \min(sd(SP),1)
$$
Where $\rho, \mu \in [0, 1]$, $\rho + \mu = 1$, $sd$ refers to the standard deviation of a set, and $SP = \{p_{ag_{u'}}(x)| u' \in Sim^\epsilon_\nu(u, x)\}$.
\end{mydef}

Note that, in order to have confidence between 0 and 1, we set an upper bound of 1 to each of the two parts. The first part of the $\rho\mu$-Confidence refers to the separation between the users for the prediction, and the higher this separation, the less confidence in the prediction. In this case, we measure the average separation between $u$ and the users in $Sim^\epsilon_\nu(u, x)$. The second part refers to the distribution of the real preferences of the similar users, hence the more these preferences differ, the lower confidence in our prediction. Here, we use the standard deviation of the preferences.

Once we have a confidence function, we can use it to define variable thresholds to create norms from preferences. One possibility is to favour the creation of norms when we have confident predictions, while limiting their production when we have low confidence. In other words, we can consider variable thresholds that are closer to the middle point (0) when confidence is high, and closer to the extremes (-1 and 1) when confidence is low. 

\begin{mydef}[Confident norm thresholds]
Given a confidence function $conf(u, x)$, we define confident norm thresholds as $\theta_{prh}(u, x) = -1 + \frac{conf(u, x)}{3}$, $\theta_{per}(u, x) = 1 - 2\frac{conf(u, x)}{3}$.
\end{mydef}

We use confident norm thresholds for our running example.

\begin{myexample}
We want to infer a norm for $u_1$ and element $x_3$, in Example \ref{ex:pred} where we predicted a preference of $-1$. Note that, if we consider $\rho=\frac{1}{2}, \mu=\frac{1}{2}$, in this case we have $conf_{\rho, \mu}(u_1, x_3) = 1$ (as both parts of the function are 0). Hence, we would have $\theta_{prh}(conf) = -\frac{2}{3}$, $\theta_{per}(conf) = \frac{1}{3}$, and would infer $Prh(x_3)$ in this case, because $-1 < \theta_{prh}$.
\end{myexample}

\subsection{Thresholds based on other variables}

Much like with prediction confidence, threshold functions can also depend on other relevant variables like the context of the elements (assuming they have contexts). For generality purposes, we have avoided defining formally any type of these variables, and have considered them implicitly in each $x \in X$. However, in some applications it might be important to consider them when setting norm thresholds. For example, if we want to avoid inappropriate actions in sensitive contexts, we can consider the sensitivity of the context as a variable to set the thresholds. Then, $\epsilon_{prh}(c)$ would be closer to 0 for contexts $c$ that are considered sensitive than for non-sensitive ones. Formally, in this case, we would consider thresholds as functions depending on multiple variables $\epsilon_{prh}: V_1 \times V_n \rightarrow [-1, 0]$ and $\epsilon_{per}: V'_1 \times V'_m \rightarrow [0, 1]$, where we consider $n$ and $m$ variables respectively and $V_1 \times V_n$ and $V'_1 \times V'_m$ are the possible values of these $n$ and $m$ variables. As for hard thresholds, we require that the $\epsilon_{prh}$ and $\epsilon_{per}$ functions have ranges in $[-1, 0]$ and $[0, 1]$ respectively.

\subsection{The suitability of the different approaches}

The suitability of each of the previous approaches depends largely on the domain of application. Apart from particular application requirements, when deciding which method to apply we should also consider the accuracy and distribution of predictions. To be concise, we discuss this in relation to two general measures: the average prediction distance from the real preference (denoted as APD), as well as the standard deviation of these predictions (denoted PSD). The average prediction distance tells us the accuracy of our predictions, while the standard deviation gives us an indication of the polarisation of predictions with regard to average distance. These two measures lead to the following four differentiated cases:

\begin{itemize}
    \item{Low APD and low PSD:} This is the ideal scenario in which predictions work best, where any method is valid. Hard thresholds are useful for cases that demand an easily explainable method. Function thresholds can also be useful, especially if required by the application (for example, one that explicitly demands consideration of environmental variables like context sensitivity when determining norms).
    \item{Low APD and high PSD:} Here, the predictions seem accurate but not reliable enough, so hard thresholds are best avoided. Instead, the other approaches prevent norms being enacted in cases of vague predictions coming from bad quality data.
    \item{High APD and low PSD:} Here the predictions are consistently wrong, and consistently deviate from the truth. This case should not usually arise and tells us that there is something wrong with the prediction formula, so predictions should not be used to build norms.
    \item{High APD and high PSD:} Here, the predictions are seemingly random. This can be a consequence of insufficient information (e.g., when known preferences are far fewer than unknown ones). At this stage, norms could be built using function thresholds, but if more information is collected this scenario should then settle into one of the other three, and norms would be selected using the relevant advice. Predictions at this stage may not be reliable so the resulting norms should be rebuilt once more information is known.
\end{itemize}
\section{Proof of concept: Privacy norms for Smart Personal Assistants}
\label{sec:case}

In order to validate the preference prediction and norm inference models presented here, we return to the problem 
of Smart Personal Assistants. We consider this case by virtue of the privacy preferences dataset used by \cite{NouraPrivacyNorms2021}, which is available at \cite{NouraDataset}.

\subsection{Description of the dataset}

The dataset contains the responses of 1737 participants in a survey concerning privacy preferences when using Smart Personal Assistants (SPAs). The questions in the survey\footnote{For our tests, we only use the main block of questions in the survey, so to not consider data from questions on demographics, IUIPC, and security attitudes} ask participants how acceptable it is to share data in a particular context. More specifically, the survey considers 15 data types (e.g. emails, banking data, healthcare data, voice recordings) and presents 8 scenarios for each. These scenarios or contexts consider different recipients of the data (e.g.\ parents, friends, visitors), the purpose of sharing the data, different conditions on data transmission, etc. Each scenario has a different number of associated questions, amounting to 55 for all 8 scenarios. Overall, the survey consists of 825 different preference questions, for each of which participants answer on a 1 to 5 Likert scale (1 meaning the sharing of that datatype is completely unacceptable in that context, and 5 meaning it is completely acceptable). Participants did not answer all questions, with each participant answering questions related to 4 scenarios for 6 datatypes (both selected randomly). Note that different scenarios have different amounts of associated questions, so participants responded to different questions and different numbers of questions, ranging from 144 to 199, with an average of 170 questions answered. 

\subsection{Prediction validation}

In this section we assess the accuracy of our predictions using the previously described dataset\marc{\footnote{The code necessary to run these experiments can be found at: \url{https://github.com/secure-ai-assistants/norm-prediction}}}. We show that our predictions are more accurate than random guesses and also more accurate than the preferences found by Abdi et al. in \cite{NouraPrivacyNorms2021}. Finally, we record the confidence measure for each prediction and show that there is a correlation between confidence and prediction quality. 

First, we describe the experiments on accuracy.
Out of the 1737 participants, we selected 20\% of participants (347) randomly to test the accuracy of our predictions. The remaining 80\% of participants (1390) represent our base of knowledge to build predictions. For each test participant, we randomly picked 20\% of their answers as the test set we sought to predict (this was an average of 35). We did not consider all remaining answers to assess user similarity, but instead used only 40\% of answers (an average of 71), and applied this reduction to all participants. 
To proceed, for each test participant and answer to predict, we filtered the pool of 1390 participants to keep only those relevant; we needed to filter out participants who did not answer the question we aim to predict. In addition, we discarded participants with less than 5 questions in common with the test participant (as we wanted to finding similar users with a certain degree of reliability). Then, we assessed the separation between users using the cumulative user separation measure of Def.~\ref{def:exsep}.
Using this separation measure, we selected participants similar to our test participant as in Def.\ \ref{def:ensimilar}, with $\epsilon = 0$, and $\nu = 5$. In other words, we selected all users at distance 0, and then selected those with the least distance until we had 5 participants. Then, we predicted the test participant's answers to the test questions using the average preference prediction function (see Def. \ref{def:avgpred}).

To test our prediction, we calculated the distance between our predicted answer and the real answer, and collected all these distances for all test participants and test questions. Below, we report the mean distance and the standard deviation. Apart from the experiment considering all participants (which we now refer to as the {\em regular} experiment) we wanted to test accuracy for two further levels of difficulty. First, we hypothesised that participants that responded similarly to all questions would be easiest to predict. Thus, when selecting test participants we avoided those with small standard deviations on their given answers (but retained them in the pool of possible similar participants). Here, we required the standard deviation of any selected test participant to be no less than 1. We repeated the experiments as explained above, and refer to this as the {\em medium hardness} experiment. Second, to test the extreme case, we selected the 100 participants with highest standard deviation in their answers and repeated the experiments as explained above. These are arguably the most difficult participants to predict. By selecting many of them, we limit the chances of finding very similar participants in the remaining pool (since, to be similar, they also need to have a large standard deviation and therefore might have been selected as test participants). We call this the {\em hard} experiment.

Table \ref{tab:disresults} provides the mean distance as well as the standard deviation for each run of tests. We provide a histogram for the regular experiments to better understand the distribution of distances between prediction and reality (the histograms for medium and hard experiments are almost identical, so we omit them to save space). 

\begin{table}[b]
\begin{tabular}{|l|l|l|l|}
\hline
 & Regular & Medium & Hard \\ \hline
Mean distance from real answer & 0.5954 & 0.6538 & 0.7480 \\ \hline
Standard deviation & 0.6757 & 0.7207 & 0.8707 \\ \hline
\end{tabular}
\caption{Results for regular, medium and hard experiments}
\label{tab:disresults}
\end{table}

\begin{figure}
    \centering
    \includegraphics[width=0.8\linewidth]{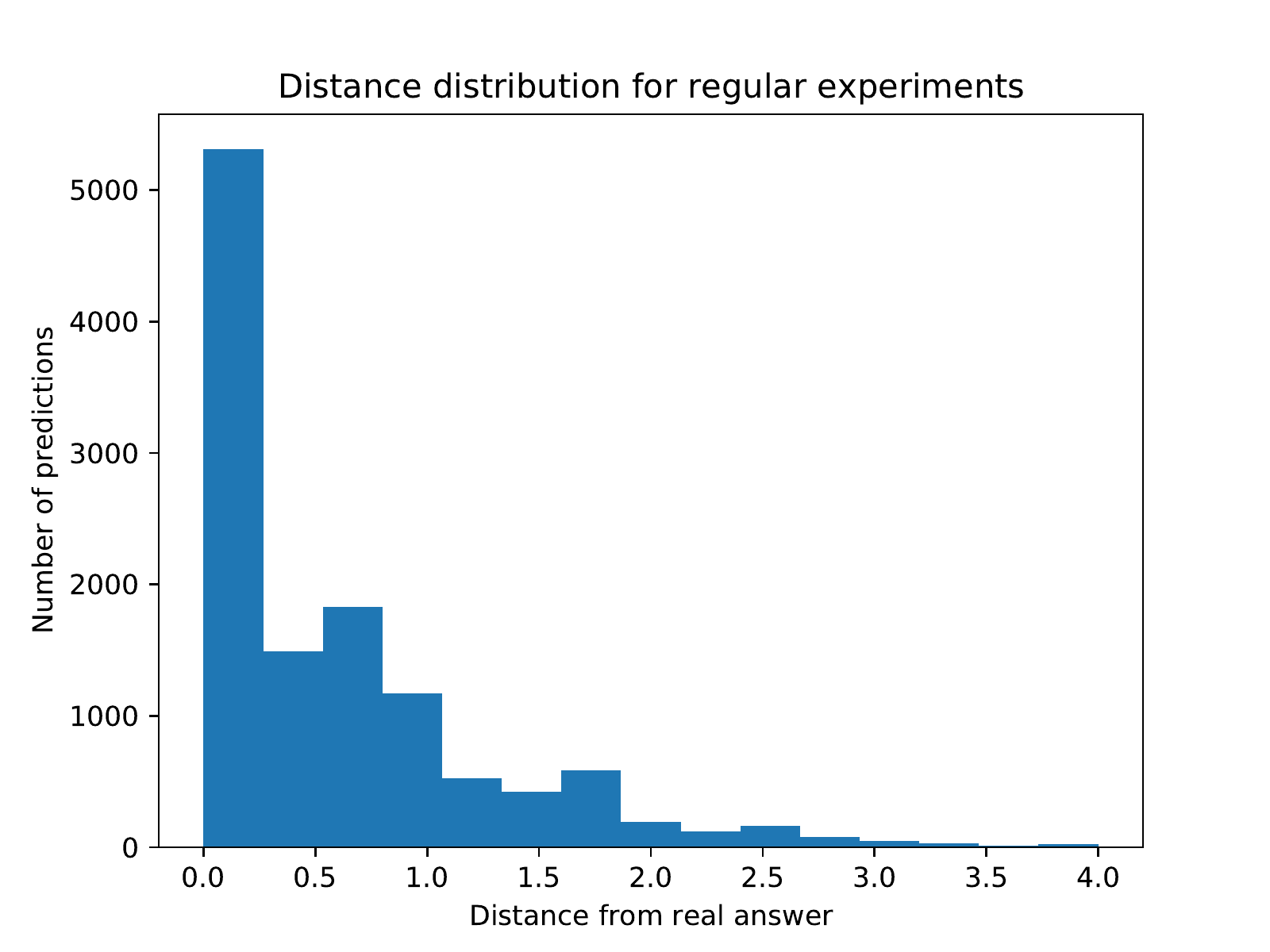}
    \caption{Histogram of distances between our predictions and the real preferences}
    \label{fig:regular}
\end{figure}

The regular experiment resulted in 12007 predicted answers with a mean distance from the real answer of 0.5954 and a standard deviation of 0.6757. Figure \ref{fig:regular} shows the distribution of these 12007 distances between prediction and reality. We can see that most predictions fall within 0.25 of the real answer, while only a small number of predictions have distances larger than 1 with regard to the real answer. The medium hardness experiment resulted in 12015 predicted answers\footnote{Different experiments have different numbers of predictions because we selected as test answers 20\% of a participants' answers and different participants answered different numbers of questions}; as expected, the mean distance grows a little to 0.6538 and the standard deviation to 0.7207. Overall, while distances grow slightly, we can see that predictions are still of good quality. 
Finally, the hard experiment resulted in 3461 predicted answers\footnote{Note that the hard experiment has fewer predicted answers because it consists of 100 test participants instead of 347} with a mean distance of 0.7480 and a standard deviation of 0.8707. While the mean distance has increased from the distance of the regular experiment it has increased by less than 0.2. The standard deviation has increased by almost 0.2, confirming that the predictions of these participants may lead to more outliers.

Thus, while increasing the difficulty of predictions slightly increases the distance between the prediction and the real answer, we have seen that our predictions are still reliable with the hardest participants to predict. Furthermore, our predictions improve the preferences presented by Abdi et al. in \cite{NouraPrivacyNorms2021}, which are the average preferences for all participants in their survey. \marc{Most notably, our model can capture preferences outside the majority view; for example, Abdi et al. point out that people are less inclined to share video call data with assistant providers (the average preference for sharing this while being able to delete it is 2.44), yet in many instances we correctly predicted favourable preferences towards sharing.} Table \ref{tab:noura} shows the results of our experiments considering their preferences, and as an additional point of reference, Table \ref{tab:random} provides results for the experiments with random predictions.%

\begin{table}[t]
\begin{tabular}{|l|l|l|l|}
\hline
 & Regular & Medium & Hard \\ \hline
Mean distance from real answer & 1.0437 & 1.0611 & 1.2895 \\ \hline
Standard deviation & 0.7069 & 0.7093 & 0.7765 \\ \hline
\end{tabular}
\caption{Results 
with preferences found by Abdi et al. \cite{NouraPrivacyNorms2021} }
\label{tab:noura}
\end{table}


\begin{table}[b]
\begin{tabular}{|l|l|l|l|}
\hline
 & Regular & Medium & Hard \\ \hline
Mean distance from real answer & 1.6083 & 1.578 & 1.8768 \\ \hline
Standard deviation & 1.0864 & 1.080 & 1.1286 \\ \hline
\end{tabular}
\caption{Results 
with random predictions}
\label{tab:random}
\end{table}

When it comes to confidence, we tested how $\rho\mu$-Confidence correlates with prediction quality (its closeness to the real preference) using Spearman's correlation coefficient\footnote{We cannot ensure that our data follows a normal distribution, hence Spearman's is the appropriate correlation test to use}. Note that the coefficient must be a negative number as we should have an inverse correlation (the higher the confidence, the lower the distance between the real and predicted preferences). While our confidence formula considers two weights $\rho$ and $\mu$, these depend on each application and could be adjusted at runtime to maximise correlation (i.e.\ to minimise the correlation coefficient). For regular experiments, we have a correlation coefficient of -0.67 with $\rho =0$ and $\mu =1$. With medium hardness experiments, the minimum was -0.63 with $\rho =0.01$ and $\mu =0.99$. Finally, for hard experiments, the minimum was -0.74 when $\rho =0.15$ and $\mu =0.85$. We can therefore detect an inverse correlation between confidence and prediction quality. We also see that in this case, confidence largely depends on the distribution of preferences from which the prediction is made, whereas the separation between the user and the similar users is not pertinent to assess confidence. Importantly, we see that our confidence function is more reliable for hard predictions. We believe this is because since confidence is always in [0,1] it is easier for it to correlate with prediction quality in cases where the quality has more variability (i.e.\ in the case of hard experiments).

\subsection{Evaluating inferred norms with real users}


To test the norm inference process with real users we performed a user study with the scenarios from \cite{NouraPrivacyNorms2021} and the preference data collected in \cite{NouraDataset}, with the aim of validating user perceptions of our predicted norms. Through Prolific,\footnote{\url{prolific.co}} we recruited 50 participants matching the demographics of the original data set who answered 32 preference questions over 5 randomly selected scenarios from \cite{NouraPrivacyNorms2021, NouraDataset}. We then selected three unknown preferences at random and made predictions for them, inferring the norms using the hard thresholds function (see Section \ref{sub:hard}); if no norm was generated for a preference we randomly selected another unknown preference\marc{\footnote{Note that this does not compromise our results as we only aim to validate the generated norms. If we do not predict a clear preference, our approach does not produce any norm and instead we resort to other approaches (like asking for consent).}}. We also interleaved three control norms with the same structure but randomly generated outcomes. Participants rated these norms using 5 point Likert items from completely inappropriate (1) to completely appropriate (5) and could leave a text comment explaining their reasoning. After discarding 3 incorrect responses to the included attention check, the remaining 47 participants had an average age of 37.5 ($\sigma=13.8$), and 49\% identified as women. The study was approved by our Institutional Review Board (IRB).

\begin{figure}
    \centering
    \includegraphics[width=0.7\linewidth]{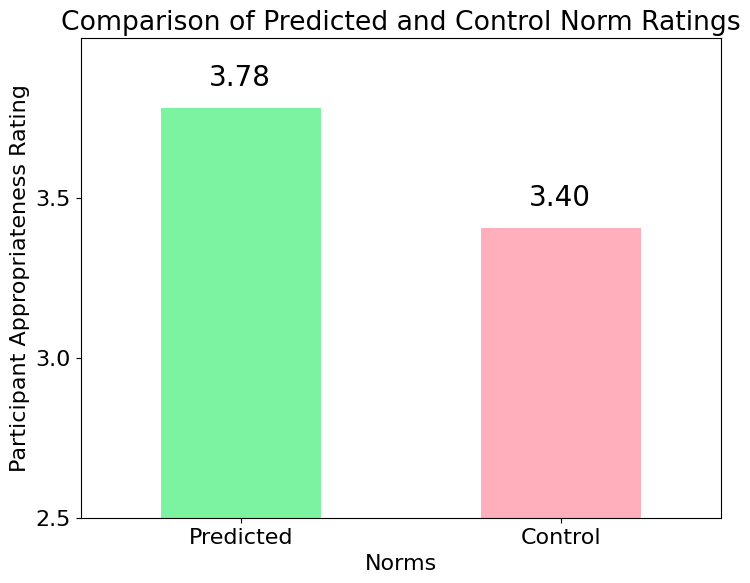}
    \caption{Comparison between mean participant ratings of predicted and control norms}
    \label{fig:validation}
\end{figure}

Figure~\ref{fig:validation} shows the mean ratings of the 141 predicted and 141 control norms, showing a substantial improvement of predicted norms over the control set. Overall we predicted 126 unique preferences covering 15\% of the original set. A follow up t-test (t=2.88, p=0.003) confirms that participants' higher agreement with our predicted norms compared to the control norms is statistically significant, with the prediction process eliminating almost a quarter (24\%) of the difference between the control ratings and the perfect acceptance score of 5. As in similar user studies (e.g., ~\cite{10.5555/3535850.3535989}), we emphasise that it is very unlikely that even ground truth preferences would receive a perfect score of 5 given the variability of self-ratings and the tendency for participants to bring in outside context when evaluating norms (as seen in accompanying text comments).



\section{Related Work}
\label{sec:rw}

Some previous approaches have addressed the problem of privacy norms in AI assistants. For example, Abdi et al.~\cite{NouraPrivacyNorms2021} surveyed users on how acceptable some information exchanges are under some context (this survey produced the data we used for our experiments \cite{NouraDataset}) and then crowd-sourced norms that aligned with their answers. While this work is useful to understand the general preferences of users towards privacy, it is restricted in terms of the scenarios covered. Nonetheless, it could be useful as a default set of preferences when information about the user is sparse (e.g. when the user first uses the assistant). The work of Zhan et al.~\cite{Xiao2022Model} instead proposes to construct AI assistant privacy norms using an approach based on rule mining and machine learning (exploiting the idea of contextual integrity \cite{nissenbaum2004privacy}). Unfortunately, we cannot compare our norm prediction approach with their approach as \cite{Xiao2022Model} reports test results as percentages of accuracy which cannot be compared with our acceptability rates. While this approach achieves an accuracy of 70-80\% it requires domain knowledge. In contrast our work in this paper can customise norms to each user without the need for specifying or formalising contexts. Note that the elements in our set $X$ can be actions, or tuples of actions and contexts, and in both cases our method is able to predict preferences and infer norms without the need for additional knowledge of the elements in $X$. 

Closely related to this is the area of AI ethics and norms. We can assume that morality influences a user's preferences towards AI so that, for example, a user that highly values privacy will be less inclined to share data. Works like \cite{serramia2018aamas, serramia2021JAAMAS} have investigated the selection of norms with regard to their promotion of moral values and preferences over these values. Similarly, \cite{sierra2019value} proposes to enact those norms that will benefit those state transitions leading to an increase in value alignment with the considered values and preferences. In this direction, Montes et al. \cite{montessynthesis22} describe how these norms can be synthesised. While these approaches could produce privacy norms (provided privacy is a value considered) we argue that in practice this would not be possible since they require knowledge that is hardly attainable by smart devices, like states of the world, contexts, the user's value preferences, or a measurement of value alignment (with regard to privacy and other desirable values).
\section{Conclusions}
\label{sec:conc}

Collaborative filtering is a useful tool in recommender systems. For example, online stores use it to recommended products by considering purchases of similar users. This paper provides a novel application of collaborative filtering, with the aim of predicting user preferences towards AI. However, our approach offers far more than just recommending preferences to the user. Indeed, while users expect smart devices to act as they desire, constant interaction not only annoys the user but fails to capture their true preferences. Hence, our approach has two purposes: understanding user preferences while minimising interaction, and bringing more value to interactions regarding preferences by considering predictions. Thus, coupling collaborative filtering with norms allows us to both add a component of explainability to user preferences, and to propagate user preferences to other parts of the AI ecosystem. For example, in the case of privacy in smart assistants, norms could govern the management of data not only by the device itself but also for other components of the ecosystem, like skills.

Admittedly, our approach requires large quantities of users and partial preferences for each of these users to function properly, and the more information the more accurate the predictions. Thus, our approach might be better suited to smart devices with a reasonable number of users. Even if the number of users is sufficient, it is also possible that predictions could be unreliable. However, we can detect this using a confidence measure, such as that of Definition \ref{def:rmconf}. Crucially, however, as the number of users grows, and as knowledge of their preferences increases, low confidence preferences can be recalculated, which should increase the confidence in the prediction.

In addition, we have assumed a single user for each device, but it is unclear how this method would apply when multiple users share the same device (for example, a family sharing a smart speaker). This will be the subject of future work. \marc{Other interesting aspects we plan to investigate include the addition of rewards or punishments associated with norms (which could be derived from context sensitivity), and how to produce explanations from norms.}


\balance



\begin{acks}
Research funded by  project SAIS Secure AI AssistantS via Grant EP/T026723/1, funded by the UK Engineering and Physical Sciences Research Council; and by project  TED2021-131295B-C32, funded by MCIN/AEI/ 10.13039/501100011033 and the European Union NextGenerationEU/PRTR.
\balance
\end{acks}



\bibliographystyle{ACM-Reference-Format} 
\bibliography{Bibliography}


\end{document}